\documentclass{article} % For LaTeX2e
\usepackage{iclr2026_conference,times}

\usepackage[utf8]{inputenc} % allow utf-8 input
\usepackage[T1]{fontenc}    % use 8-bit T1 fonts
\usepackage{hyperref}       % hyperlinks
\usepackage{nccmath}
\usepackage{pgfplots}
\pgfplotsset{compat=1.18}
\hypersetup{
     colorlinks=true,
     linkcolor=blue,
     filecolor=blue,
     citecolor=blue,
     urlcolor=blue,
     }
\usepackage{url}
\usepackage{upgreek}
\usepackage{wrapfig}
\usepackage{booktabs}       % professional-quality tables
\usepackage{amsfonts}       % blackboard math symbols
\usepackage{nicefrac}
\usepackage{microtype}      % microtypography
\usepackage{xcolor}         % colors
\usepackage{multicol}
\usepackage{graphicx}
\usepackage{amsthm}
\usepackage{bbm}
\usepackage{comment}
\usepackage{enumitem}
\usepackage{bm}
\usepackage{booktabs}
\usepackage{subfigure}
\usepackage{mathtools}
\usepackage[export]{adjustbox}
\usepackage{stmaryrd}
\usepackage{booktabs} % for professional tables
\usepackage{xifthen}
\usepackage{xargs}
\usepackage{array}
\usepackage{setspace}
\usepackage{mathtools}
\usepackage{amsmath,amssymb}
\usepackage{algorithm,algorithmic}
\usepackage{caption}
\usepackage{cleveref}
\usepackage{titlesec}
\usepackage{multirow}
\usepackage{subfigure}
\usepackage{mathrsfs}
\usepackage{colortbl}
\usepackage[utf8]{inputenc}
\usepackage{pgfplots}
\DeclareUnicodeCharacter{2212}{−}
\usepgfplotslibrary{groupplots,dateplot}
\usetikzlibrary{patterns,shapes.arrows}
\pgfplotsset{compat=newest}

\titleformat*{\subparagraph}{\itshape}

\numberwithin{equation}{section}

\definecolor{babypink}{rgb}{0.96, 0.76, 0.76}
\definecolor{burntsienna}{rgb}{0.91, 0.45, 0.32}     % colors
\definecolor{crimson}{rgb}{0.86, 0.08, 0.24}
\definecolor{darkspringgreen}{rgb}{0.09, 0.45, 0.27}
\definecolor{highlighColor}{rgb}{0.91, 0.41, 0.17}  % orange
\definecolor{royalblue}{rgb}{0.25, 0.41, 0.88}
\definecolor{royalpurple}{rgb}{0.47, 0.32, 0.66}
\definecolor{ruddy}{rgb}{1.0, 0.0, 0.16}
\definecolor{deepcarrotorange}{rgb}{0.91, 0.41, 0.17}
\definecolor{darkspringgreen}{rgb}{0.09, 0.45, 0.27}

\definecolor{rebuttalColor}{rgb}{0.17,0.63,0.17}
\usepackage{transparent}

\usepackage[textsize=tiny,textwidth=1.15\linewidth]{todonotes}
\makeatletter
\tikzset{
    notestyleraw/.append style={
        fill opacity=0.8,    % Adjusts the background transparency (0.0 to 1.0)
        text opacity=1       % Keeps the text fully opaque for readability
    }
}
\makeatother
\def\Id{\mathrm{I}}

\def\pE{\mathbb{E}}

\def\wrt{w.r.t.}

\newcommandx\targ[4][4=]{
    \ifthenelse{\equal{#2}{}}{
        \ifthenelse{\equal{#3}{}}{
            p^{#4} _{#1}
        }{
            p^{#4} _{#1}(#3)
            }
    }{
        \ifthenelse{\equal{#3}{}}{
            p^{#4} _{#1}(\cdot|#2)
        }{
            p^{#4} _{#1}(#3|#2)
        }
    }}

\newcommandx\interp[4][4=\interpscale]{
    \ifthenelse{\equal{#2}{}}{
        \ifthenelse{\equal{#3}{}}{
            \pi^{#4} _{#1}
        }{
            \pi^{#4} _{#1}(#3)
            }
    }{
        \ifthenelse{\equal{#3}{}}{
            \pi^{#4} _{#1}(\cdot|#2)
        }{
            \pi^{#4} _{#1}(#3|#2)
        }
    }}

\newcommandx\pdata[4][4=]{
    \ifthenelse{\equal{#2}{}}{
        \ifthenelse{\equal{#3}{}}{
            p^{#4} _{#1}
        }{
            p^{#4} _{#1}(#3)
        }
    }{
        \ifthenelse{\equal{#3}{}}{
            p^{#4} _{#1}(\cdot|#2)
        }{
            p^{#4} _{#1}(#3|#2)
        }
    }}

\newcommandx\hpdata[4][4=]{
    \ifthenelse{\equal{#2}{}}{
        \ifthenelse{\equal{#3}{}}{
            \hat{p}^{#4} _{#1}
        }{
            \hat{p}^{#4} _{#1}(#3)
        }
    }{
        \ifthenelse{\equal{#3}{}}{
            \hat{p}^{#4} _{#1}(\cdot|#2)
        }{
            \hat{p}^{#4} _{#1}(#3|#2)
        }
    }}

\newcommandx\revker[4][4=]{
\ifthenelse{\equal{#2}{}}{
    \ifthenelse{\equal{#3}{}}{
        r^{#4} _{#1}
    }{
        r^{#4} _{#1}(#3)
    }
}{
    \ifthenelse{\equal{#3}{}}{
        r^{#4} _{#1}(\cdot|#2)
    }{
        r^{#4} _{#1}(#3|#2)
    }
}}

\newcommandx\hatrevker[4][4=]{
    \ifthenelse{\equal{#2}{}}{
        \ifthenelse{\equal{#3}{}}{
            \hat{r}^{#4} _{#1}
        }{
            \hat{r}^{#4} _{#1}(#3)
        }
    }{
        \ifthenelse{\equal{#3}{}}{
            \hat{r}^{#4} _{#1}(\cdot|#2)
        }{
            \hat{r}^{#4} _{#1}(#3|#2)
        }
    }
}

\newcommandx\refreshker[4][4=]{
    \ifthenelse{\equal{#2}{}}{
        \ifthenelse{\equal{#3}{}}{
            m^{#4} _{#1}
        }{
            m^{#4} _{#1}(#3)
        }
    }{
        \ifthenelse{\equal{#3}{}}{
            m^{#4} _{#1}(\cdot|#2)
        }{
            m^{#4} _{#1}(#3|#2)
        }
    }}
\newcommand\jpdata[3]{
    \ifthenelse{\equal{#2}{}}{
        \ifthenelse{\equal{#3}{}}{
            \bar{p}_{#1}
        }{
            \bar{p}_{#1}(#3)
        }
    }{
        \ifthenelse{\equal{#3}{}}{
            \bar{p}_{#1}(\cdot|#2)
        }{
            \bar{p}_{#1}(#3|#2)
        }
    }}

\newcommandx\post[4][4=]{
    \ifthenelse{\equal{#2}{}}{
        \ifthenelse{\equal{#3}{}}{
            \pi^{#4} _{#1}
        }{
            \pi^{#4} _{#1}(#3)
        }
    }{
        \ifthenelse{\equal{#3}{}}{
            \pi^{#4} _{#1}(\cdot|#2)
        }{
            \pi^{#4} _{#1}(#3|#2)
        }
    }}

\newcommandx\hpost[4][4=]{
    \ifthenelse{\equal{#2}{}}{
        \ifthenelse{\equal{#3}{}}{
            \hat\pi^{#4} _{#1}
        }{
            \hat\pi^{#4} _{#1}(#3)
        }
    }{
        \ifthenelse{\equal{#3}{}}{
            \hat\pi^{#4} _{#1}(\cdot|#2)
        }{
            \hat\pi^{#4} _{#1}(#3|#2)
        }
    }}

  \newcommandx\pot[3][3=]{
        \ifthenelse{\equal{#3}{}}{
            \ell^{#3} _{#1}(\obs|#2)
        }{
            \ell^{#3} _{#1}(\obs|#2)
        }
    }

\newcommandx\hpot[4][4=]{
    \ifthenelse{\equal{#2}{}}{
        \ifthenelse{\equal{#3}{}}{
            \hat{\ell}^{#4} _{#1}
        }{
            \hat{\ell}^{#4} _{#1}(#3)
        }
    }{
        \ifthenelse{\equal{#3}{}}{
            \hat{\ell}^{#4} _{#1}(\cdot|#2)
        }{
            \hat{\ell}^{#4} _{#1}(#3|#2)
        }
    }}

\newcommandx\fw[4][4=]{
        \ifthenelse{\equal{#3}{}}{
            q^{#4} _{\smash{#1}}(\cdot|#2)
        }{
            q^{#4} _{\smash{#1}}(#3|#2)
        }
    }

\newcommandx\denoiser[5][4=, 5=0]{
    \ifthenelse{\equal{#2}{}}{
        \ifthenelse{\equal{#3}{}}{
            \hat\bx^{#4}_{#5}(\cdot, #1)
        }{
            \hat\bx^{#4} _{#5}(#3, #1)
        }
    }{
        \ifthenelse{\equal{#3}{}}{
            \hat\bx^{#4} _{#5}(\cdot, #1|#2)
        }{
            \hat\bx^{#4} _{#5}(#3, #1|#2)
        }
    }
}

\newcommandx\noisepred[4][4=]{
    \ifthenelse{\equal{#2}{}}{
        \ifthenelse{\equal{#3}{}}{
            \hat\bx^{#4}_{1}(\cdot, #1)
        }{
            \hat\bx^{#4} _{1}(#3, #1)
        }
    }{
        \ifthenelse{\equal{#3}{}}{
            \hat\bx^{#4} _{1}(\cdot, #1|#2)
        }{
            \hat\bx^{#4} _{1}(#3, #1|#2)
        }
    }
}

    \newcommandx\hdenoiser[4][4=]{
    \ifthenelse{\equal{#2}{}}{
        \ifthenelse{\equal{#3}{}}{
            \hat{D}^{#4}_{#1}
        }{
            \hat{D}^{#4} _{#1}(#3)
        }
    }{
        \ifthenelse{\equal{#3}{}}{
            \hat{D}^{#4} _{#1}(\cdot|#2)
        }{
            \hat{D}^{#4} _{#1}(#3|#2)
        }
    }}

\newcommandx\epspred[4][4=]{
    \ifthenelse{\equal{#2}{}}{
        \ifthenelse{\equal{#3}{}}{
            \varepsilon^{#4}_{#1}
        }{
            \varepsilon^{#4} _{#1}(#3)
        }
    }{
        \ifthenelse{\equal{#3}{}}{
            \varepsilon^{#4} _{#1}(\cdot|#2)
        }{
            \varepsilon^{#4} _{#1}(#3|#2)
        }
    }}

\newcommand\clf[3]{
    \ifthenelse{\equal{#2}{}}{
        g_{#1}(#3|\cdot)
    }{
        g _{#1}(#3|#2)
    }
}

\newcommand\cfgdist[3]{
    \ifthenelse{\equal{#2}{}}{
        p^w _{#1}(#3|\cdot)
    }{
        p^w _{#1}(#3|#2)
    }
}

\newcommand\cscore[3]{
    \ifthenelse{\equal{#2}{}}{
        \ifthenelse{\equal{#3}{}}{
            s _{#1}
        }{
            s _{#1}(#3)
        }
    }{
        \ifthenelse{\equal{#3}{}}{
            s _{#1}(\cdot|#2)
        }{
            s _{#1}(#3|#2)
        }
    }}

\def\gauss{\mathcal{N}}

\def\interpscale{\eta}

\def\bx{\mathbf{x}}

\def\blended{{\sc{Blended-Diff}}}

\def\ding{{\sc{DInG}}}
\def\ddnm{{\sc{DDNM}}}
\def\diffpir{{\sc{DiffPIR}}}

\def\flowchef{{\sc{FlowChef}}}

\def\flair{{\sc{FLAIR}}}

\def\div2k{{\texttt{DIV2K}}}

\def\sd3{{{SD3}}}
\def\flux{{{FLUX}}}
\def\ltx{{{LTX}}}
\def\wan{{{Wan2.1}}}

\def\param{\theta}

\def\obs{\mathbf{y}}

\newcommandx{\hpredx}[3][2=0,3=\param]{\smash{m^{#3} _{#2|#1}}}
\newcommandx{\predx}[2][2=0]{\smash{m _{#2|#1}}}
\newcommandx{\prednoise}[2][2=\param]{\smash{\epsilon^{#2} _{#1}}}
\newcommandx{\score}[2][2=\param]{s^{#2} _{#1}}

\def\inpaintcoco{{\texttt{InpaintCOCO}}}
\def\humanedit{{\texttt{HumanEdit}}}
\def\vpbench{{\texttt{VPBench}}}

\definecolor{BaseHighlight}{HTML}{1f77b4}

\def\runtimecolor{red}

\def\packageURL{https://github.com/Badr-MOUFAD/ding-editor}
\def\webpageURL{https://badr-moufad.github.io/ding-editor/}
\def\packageName{\texttt{DInG-editor}}

\newcommand\blfootnote[1]{%
  \begingroup
  \renewcommand\thefootnote{}\footnote{#1}%
  \addtocounter{footnote}{-1}%
  \endgroup
}

\newcounter{hypA}

\title{When Test-Time Guidance Is Enough: Fast Image and Video Editing with Diffusion Guidance}

\iclrfinalcopy

\author{% 
Ahmed Ghorbel\textsuperscript{1,*} \,
Badr Moufad\textsuperscript{1,*} \,
Navid Bagheri Shouraki\textsuperscript{1,5,7} \,
Alain Oliviero Durmus\textsuperscript{1} \,
\\
{\bf{%
Thomas Hirtz\textsuperscript{6} \,
Eric Moulines\textsuperscript{3,4} \,
Jimmy Olsson\textsuperscript{8} \,
Yazid Janati\textsuperscript{2,3,*} \,
}}
\\
\\
\textsuperscript{1}CMAP, Ecole Polytechnique \,
\textsuperscript{2}Institute of Foundation Models \,
\textsuperscript{3}MBZUAI \,
\textsuperscript{4}EPITA \,
\\
\textsuperscript{5}Sorbonne University \,
\textsuperscript{6}Lagrange Mathematics and Computing Research Center \,
\\
\textsuperscript{7}EPITA Research Lab \,
\textsuperscript{8}KTH Royal Institute of Technology \,
}

\begin{document}
\maketitle

\begin{abstract}
  Text-driven image and video editing can be naturally cast as inpainting problems, where masked regions are reconstructed to remain consistent with both the observed content and the editing prompt. Recent advances in test-time guidance for diffusion and flow models provide a principled framework for this task; however, existing methods rely on costly vector--Jacobian product (VJP) computations to approximate the intractable guidance term, limiting their practical applicability. Building upon the recent work of \citet{moufad2025ding}, we provide theoretical insights into their VJP-free approximation and substantially extend their empirical evaluation to large-scale image and video editing benchmarks. Our results demonstrate that test-time guidance alone can achieve performance comparable to, and in some cases surpass, training-based methods.
\end{abstract}
\blfootnote{* {Correspondence: {\{ahmed.ghorbel, \ badr.moufad\}@polytechnique.edu}, \, {yazid.janati@mbzuai.ac.ae}}}
% Text-driven image and video editing can be naturally formulated as inpainting problems, where masked regions are reconstructed in a manner consistent with both the observed content and the editing prompt.
% While recent advances in test-time guidance of diffusion and flow models can tackle this problem,
% existing methods rely on expensive vector--Jacobian product (VJP) computations to approximate the intractable guidance term, which undermines the practicality of this approach.
% We build upon the recent work of \citet{moufad2025ding}.
% We provide theoretical insights on their VJP-free approximation.
% We further extend their benchmarks conduct to more image and video editing tasks with large scale diffusion models and showcase that test-time guidance alone can achieve performance comparable to, and in some cases surpass, training-based approaches.
% We provide theoretical insights the derivation this approximation.
% that decouples the guidance computation from the model, yielding cheap and closed-form posterior updates for linear inverse problems.
% Our results establish VJP-free test-time guidance as competitive paradigm for high-quality image and video editing using only black-box access to a frozen text-conditional diffusion model.

% --- section
\section{Introduction}
\label{sec:intro}

Image and video editing plays a central role in a wide range of applications, including content creation and interactive design. In the era of text-driven generative models, editing tasks can be naturally formalized as \emph{inpainting problems}, where regions of interest to be modified are masked and subsequently refilled with the desired content by manipulating a text prompt.
Inpainting is a classical problem in computer vision that has motivated extensive research, with particularly prominent advances arising from recent large-scale generative models based on diffusion and flow matching models \citep{esser2024scaling, batifol2025flux, wu2025qwen}. A direct approach to address inpainting problems is to \emph{train} or \emph{fine-tune} conditional diffusion models to explicitly incorporate additional inputs, such as masks and observed regions, to approximate the corresponding conditional distribution. While effective, these approaches incur non-negligible computational and data costs, which may be prohibitive in many practical scenarios.

An appealing alternative is \emph{test-time guidance}, which formulates inpainting as a Bayesian inverse problem: a pre-trained text-conditional diffusion model defines the prior, while a likelihood enforces consistency with observed regions. Sampling from the resulting posterior yields the desired completion without updating model weights. Despite its flexibility and strong empirical performance \citep{daras2024survey}, a central challenge lies in approximating the intractable guidance term. Existing methods use the likelihood evaluated at the model's output, which during sampling, entails repeated vector--Jacobian product computations through the model, which are expensive and limit scalability.
%
% An appealing alternative is \emph{zero-shot} or \emph{test-time guidance} approach: inpainting is formulated as a Bayesian inverse problem where a pre-trained text-conditional diffusion model serves as a prior whereas the likelihood term enforces consistency with the observed regions.
% The desired completion is obtained by sampling from the resulting posterior distribution
% while the weights of the model remain frozen.
% This approach has recently attracted significant attention due to its flexibility and strong empirical performance across a range of inverse problems \citep{daras2024survey}.
% %
% A key challenge in zero-shot posterior sampling lies in approximating the guidance term required to steer the generative process toward the posterior distribution. Since the \emph{oracle} guidance is typically intractable, practical methods rely on approximations that evaluate the gradient of the likelihood at the model's output. Such approximations require repeated Vector-Jacobian Product (VJP) computations through the model, which are computationally expensive and thus undermine the practicality of this approach.

In this work, we build upon the recently introduced VJP-free approximation proposed in \citet{moufad2025ding}. The core idea is to approximate the oracle guidance term using a mixture in which the intermediate variable of interest is decoupled from the model, thereby eliminating the need for VJP computations. This approximation yields cheap closed-form posterior updates for linear inverse problems, including inpainting tasks.
% Beyond inpainting tasks, (i) we provide theoretical insight on this approximation, along with ablation studies on a variety of linear inverse problems, shedding light on its accuracy and limitations.
% Besides, (ii) we extend the benchmarks in \citet{moufad2025ding} to video editing tasks and more image experiments, and demonstrate that \emph{test-time guidance alone} can achieve performance comparable to, and in some cases surpass, training- and fine-tuning-based approaches.
% Finually, (iii) we release a Python package \footnote{\url{https:\\github.com/gen-tii/ding}} adapted to the task fo editing via inpainting that accommodate new pre-trained models as well as training-free samplers in a drop-in replacement fashion.
Our contributions are threefold.
(i) We shed light on this new approximation and provide theoretical insights on its derivation.
(ii) We extend the benchmarks in \citet{moufad2025ding} to more image settings as well as video editing tasks, and demonstrate that \emph{test-time guidance alone} can achieve performance comparable to, and in some cases surpass, training-based approaches.
(iii) We release an open-source Python package\footnote{Link to the code of \packageName\ \url{\packageURL}} tailored to editing via inpainting, designed to easily accommodate new pre-trained models and training-free samplers.
Finally, our results establish test-time guidance as a compelling alternative for image and video editing that requires only black-box access to a pre-trained diffusion model with frozen weights.

% Our results establish test-time guidance as a compelling alternative for image and video editing that is competitive in quality, efficient, and requires only black-box access to a pre-trained text-conditional diffusion model with frozen weights.

% Besides, (ii) we conduct extensive benchmarks on image and video editing tasks, and demonstrate that \emph{test-time guidance alone} can achieve performance comparable to, and in some cases surpass, training- and fine-tuning-based approaches.

\section{Editing via Inpaiting with Diffusion Priors}
\label{sec:method}

\begin{figure}
    \centering
    \vspace*{-5mm}
    \includegraphics[width=\linewidth]{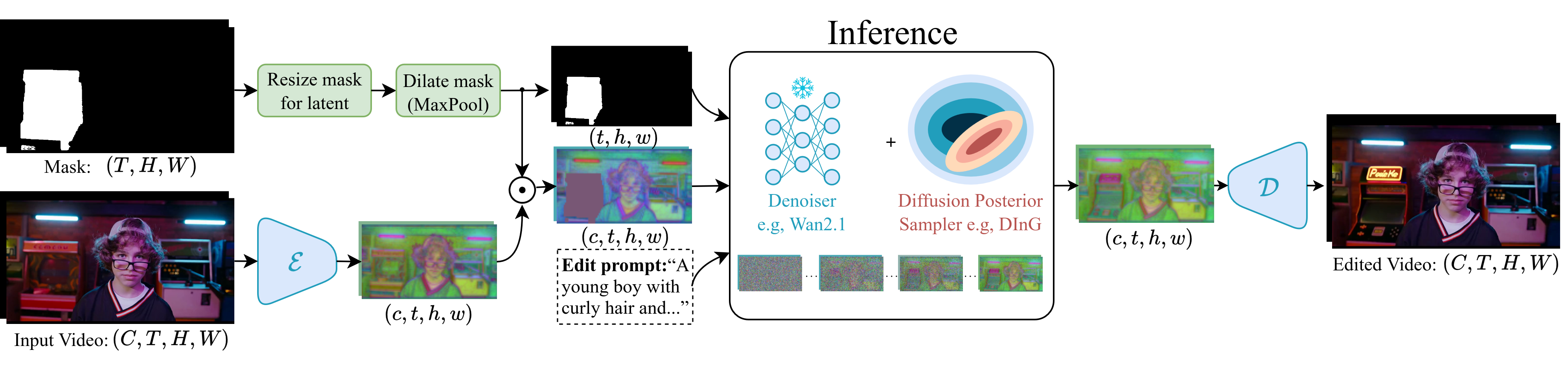}
    \captionsetup{font=small}
    \vspace*{-7mm}
    \caption{Overview of the editing pipeline for video modalities. The input video and mask are lifted to the latent space for inpainting. A pre-trained and frozen diffusion model is used with a posterior sampler to guide the generation toward prompt-aligned reconstructions, which is then decoded back to pixel space.}
    \label{fig:ding-inference-pipeline}
    \vspace*{-2mm}
\end{figure}

% \badr{to avoid ambiguity, say that diffusion and flow models are the ``facets of the same coin'' and hence we use diffusion models to refer to both.}

\paragraph{Inpainting as a Bayesian inverse problem.}
Editing tasks can be framed as inpainting problems, in which regions to be modified are masked and subsequently refilled with new content.
Let $x_\star$ denote the reference sample to be edited, and let $\mathbf{M}$ be a binary mask indicating the \emph{observed} regions to be preserved $y = \mathbf{M} \odot x_\star$.
% where $\odot$ denotes the element-wise product.
% This observation model induces the Gaussian likelihood
This induces the Gaussian likelihood
\begin{equation}
\label{eq:inpainting-prob}
\ell(y \mid x_0) % = \normpdf\!\left(y;\, \mathbf{M} \odot x_0,\, \gamma^2 \Id \right),
\propto \exp\{-\tfrac{1}{2{\gamma^{2}}}\|y - \mathbf{M} \odot x_0 \|^2\},
\end{equation}
where the parameter $\gamma > 0$ promotes stricter consistency with the observed regions, the smaller it is.
Since the likehood is agnostic to the values of masked regions, the inpainting problem is inherently ill-posed and the Bayesian formulation works around this by imposing a prior distribution $p_0$ over plausible reconstructions $x_0$.
Filling the masked regions then reduces to sampling the posterior%distribution
$$
\pi_0(x_0 \mid y) \propto \ell(y \mid x_0)\, p_0(x_0).
$$
When the prior $p_0$ is a text-conditional generative model, manipulating the prompt provides a principled way to control the semantics of the refilled regions.

\emph{Diffusion Models.}\quad
A central paradigm in modern generative modeling is learning a sequence of transformations that maps a simple reference distribution $p_1$, typically a standard Gaussian $\gauss(0,\Id)$, to a complex data distribution $p_0$. Among the many possible constructions of such transformations \citep{rezende2015variational,chen2018neural-ode,lipman2023flow,albergo2023stochastic}, diffusion models adopt the linear interpolation $X_t = \alpha_t X_0 + \sigma_t X_1$ as a principled inductive bias, where $X_0 \sim p_0$, $X_1 \sim p_1$, and $(\alpha_t,\sigma_t)$ are deterministic schedules satisfying appropriate boundary conditions. This interpolation induces a family of marginal distributions $(p_t)_{t \in [0,1]}$ that gradually transforms the data distribution $p_0$ into the Gaussian $p_1$.
Given a discretization $t_0,\ldots,t_K$ of the interval $[0,1]$, sampling from $p_0$ is performed by simulating a time-reversed Markov chain: for two consecutive timesteps $(s,t)=(t_k,t_{k+1})$, the following reverse transition is sequentially sampled
\begin{equation}
% \vspace*{-0.5mm}  
\label{eq:sampling-transition}
p^\eta_{s|t}(x_s \mid x_t) = \pE\!\left[q_{s|0,1}^\eta(x_s \mid X_0, X_1) \,\middle|\, X_t = x_t \right],
% \vspace*{-0.5mm}
\end{equation}
where the kernel $q_{s|0,1}^\eta(\cdot \mid X_0, X_1)$ preserves the path marginals
$
p_s(x_s) = \pE[q_{s|0,1}^\eta(x_s \mid X_0, X_1)],
$
whereas $\eta$ controls the stochasticity of the transition \citep{song2021ddim}.
The transition \eqref{eq:sampling-transition}
% $p^\eta_{s,t}(\cdot \mid x_t)$
is typically intractable as it depends on the unknown conditional distributions $p_{0|t}$ and $p_{1|t}$. A standard approximation consists of replacing $X_0$ and $X_1$ by their conditional expectations given $X_t$ and learning a denoiser $(t,x_t) \mapsto \hat{\mathbf{x}}_0(x_t,t)$ to approximate $\pE[X_0 \mid X_t = x_t]$ whereas the expression of the noise predictor $\hat{\mathbf{x}}_1(x_t,t)$ that approximates $\pE[X_1 \mid X_t = x_t]$ follows from Tweedie's formula \citep{roberts:tweedie:1996},
$
\hat{\mathbf{x}}_1(x_t,t) = \tfrac{1}{\sigma_t}\big(x_t - \alpha_t \hat{\mathbf{x}}_0(x_t,t)\big)%/\sigma_t
.
$
The neural network $\hat{\mathbf{x}}_0$ is then trained by minimizing a simple regression objective \citep{ho2020denoising,song2021score,karras2022elucidating}.

\begin{figure}[t]
    \centering
    % \vspace*{-5mm}
    \includegraphics[width=\linewidth]{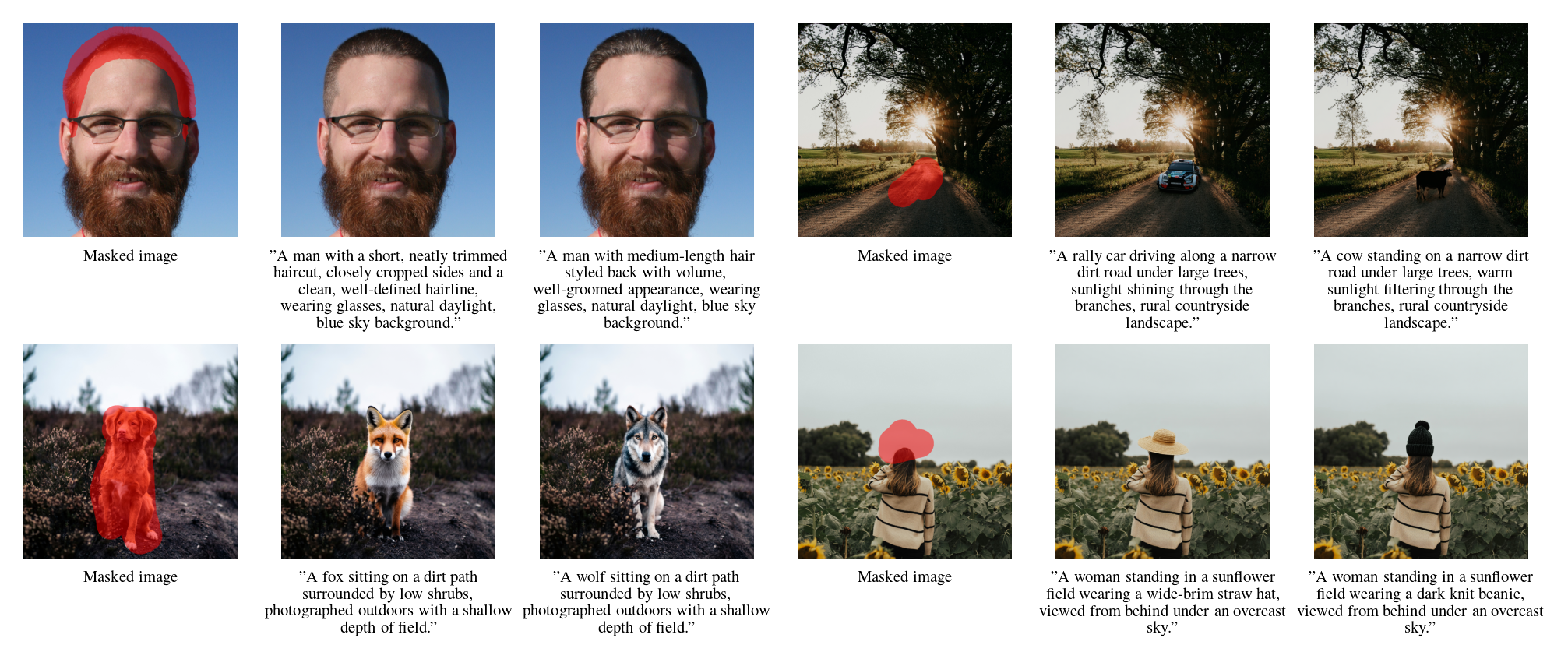}
    \captionsetup{font=small}
    \vspace*{-7mm}
    \caption{Editing via inpainting using \ding\ with \sd3\ as prior.
    Given masked inputs, the model fills the missing regions according to diverse textual prompts.
    The runtime is limited to 10 seconds per image (1024px).}
    \label{fig:examples-editing}
    % \vspace*{-4mm}
\end{figure}

\paragraph{Test-time guidance with diffusion priors.}
As sampling in diffusion models proceeds sequentially via a reverse-time Markov chain, it allows to intervene in-between to bias the transitions in \eqref{eq:sampling-transition} so as to target the posterior distribution $\pi_0(\cdot \mid y)$.
Seminal works of \citet{song2019generative,kadkhodaie2020solving,kawar2021snips} demonstrated that this can be achieved in inference time without additional training.
\citet{janati2025mgdm} derived the expression of the \emph{oracle reverse transitions}
\begin{gather}
\label{eq:posterior-transition}
\pi_{s \mid t}^{\eta}(x_s \mid x_t, y)
\propto
\ell_s(y \mid x_s)\, p^{\eta}_{s \mid t}(x_s \mid x_t),
\;\;\text{ where,}\;\;
\ell_s(y \mid x_s)
=
\mathbb{E}\!\left[\, \ell(y \mid X_0) \,\middle|\, X_s = x_s \right].
\end{gather}
The intermediate likelihood $\ell_s(y \mid x_s)$ is typically intractable as it involves the conditional $p_{0|s}(\cdot \mid x_s)$.
Applying Tweedie's formula yields the \emph{oracle conditional expectation given the observation},
\begin{equation}
\label{eq:posterior-denoiser}
\pE[X_0 \mid X_t = x_t, Y=y]
=
\pE[X_0 \mid X_t = x_t]
+
\alpha_t^{-1} \sigma_t^2 \, \nabla_{x_t} \log \ell_t(y \mid x_t),
\end{equation}
which decomposes into two terms: a prior term approximated by the pre-trained prior model $\hat{\mathbf{x}}_0(x_t,t)$, and a second term referred to as the \emph{guidance term}.
A widely used approximation introduced in \citet{chung2023diffusion} replaces the intermediate likelihood by a point estimate obtained from the denoiser output, resulting in the tractable approximation
$
\ell_t(y \mid x_t) \approx \ell\big(y \mid \hat{\mathbf{x}}_0(x_t,t)\big).
$

\paragraph{The VJP-free approximation in \citet{moufad2025ding}.}
% \paragraph{VJP-free guidance approximation.}
When combined with \Cref{eq:posterior-denoiser}, the approximation of \citet{chung2023diffusion} requires taking a gradient \wrt\ the input of the model, which entails computing a VJP. This becomes a major computational bottleneck, namely, for modern large-scale diffusion models, where VJPs are expensive.
To overcome this limitation, \citet{moufad2025ding}, building on earlier works on VJP-free approximations \citep{wang2023zeroshot,zhu2023denoising,mardani2024a}, propose a principled approximation that relies on rewriting the denoiser using the noise predictor as
$
\hat{\mathbf{x}}_0(x_s,s) = \tfrac{1}{\alpha_s}\big(x_s - \sigma_s \hat{\mathbf{x}}_1(x_s,s)\big),
$
and then replacing the input of $\hat{\mathbf{x}}_1$ by an auxiliary random variable $Z_s$ that has the same marginal as $X_s$. This yields a mixture approximation
\begin{equation}
\label{eq:ding-approx}
\vspace*{-0.25mm}
\hat{\ell}_s(y \mid x_s)
= \pE\!\left[\ell\!\left(y \mid \tfrac{1}{\alpha_s}\big(x_s - \sigma_s \hat{\mathbf{x}}_1(Z_s,s)\big)\right)\right],
\qquad
Z_s \sim p^{\eta}_{s \mid t}(\cdot \mid x_t),
\vspace*{-0.25mm}
\end{equation}
which decouples $x_s$ from the model.
Based on it, \citet{moufad2025ding} directly simulate the posterior transition~\eqref{eq:posterior-transition} by first sampling $z_s$
% $z_s \sim p^{\eta}_{s \mid t}(\cdot \mid x_t)$
and then drawing $x_s$ from the distribution $\hat{\ell}_s(y \mid \cdot, z_s)\, p^{\eta}_{s \mid t}(\cdot \mid x_t)$.
For linear inverse problems, such as inpainting~\eqref{eq:inpainting-prob}, the likelihood is Gaussian and linear in $x_s$, so the second step can be sampled exactly via Gaussian conjugacy \citep{bishop2006pattern}.
% admits a closed-form  and can be simulated exactly.

\emph{Theoretical insight.}\quad
By adding and subtracting $Z_s$ in \cref{eq:ding-approx}, the approximation rewrites as
\begin{equation*}
\vspace*{-0.25mm}
\hat{\ell}_s(y \mid x_s)
= \pE\!\left[\ell\!\left(y \mid \hat{\mathbf{x}}_0(Z_s,s) + \tfrac{1}{\alpha_s}(x_s - Z_s)\right)\right].
\vspace*{-0.25mm}
\end{equation*}
This expression can be interpreted as a first-order Taylor expansion of the denoiser $\hat{\mathbf{x}}_0(\cdot,s)$ around $Z_s$, where the true Jacobian $\nabla \hat{\mathbf{x}}_0(\cdot,s)$ is approximated by the scaled identity $(1/\alpha_s)\Id$. Differentiating the relation between the denoiser and the noise predictor yields
$
\nabla \hat{\mathbf{x}}_0(Z_s,s)
= \frac{1}{\alpha_s}\!\left(\Id - \sigma_s \nabla \hat{\mathbf{x}}_1(Z_s,s)\right).
$
Hence, the approximation in \citet{moufad2025ding} is equivalent to neglecting the Jacobian of the noise predictor.
Likewise, the second-order Tweedie formula \citep{boys2023tweedie}, i.e.
$
\nabla \mathbf{x}_0^\theta(Z_s,s)
= (\alpha_s/\sigma_s^2)\,\mathbb{C}\mathrm{ov}[X_0 \mid Z_s],
$
shows that the same approximation corresponds to assuming the that conditional covariance is isotropic
$
\mathbb{C}\mathrm{ov}[X_0 \mid Z_s] = (\sigma_s^2/\alpha_s^2)\Id.
$

\section{Experiments}

\paragraph{From pixel- to latent-space inpainting.}
\begin{wraptable}[14]{r}{0.35\textwidth}
\centering
\vspace*{-6mm}
\captionsetup{font=small}
\caption{Quantitative comparison between training-based methods and \ding\ on image editing tasks. Runtime is specified above the baselines.}
\vspace*{-2mm}
\resizebox{0.35\textwidth}{!}{
\begin{tabular}{l cccc}
\toprule
& FID & pFID & edFID & cPSNR \\
\cmidrule(lr){2-5}
\color{\runtimecolor}{30 second / image} &  \multicolumn{4}{c}{\humanedit\ 1024px} \\
\midrule
\flux + ControlNet & 26.8 & 4.0 & 24.4 & 36.5  \\
\flux\ Fill         & 27.9 & 4.0 & 22.5 & 34.1  \\
\flux + \ding       & 31.1 & 4.9 & 28.1 & 35.9  \\
\sd3 + ControlNet  & 37.1 & 11.3 & 30.6 & 26.3 \\
\sd3 + \ding        & 30.3 & 4.4 & 25.4 & 35.2  \\
\midrule
\\
\color{\runtimecolor}{10 second / image} &  \multicolumn{4}{c}{\inpaintcoco\ 512px} \\
% \cmidrule(lr){2-5}
\midrule
\flux + ControlNet & 43.1 & 13.6 & 26.8 & 31.1 \\
\flux\ Fill         & 41.0 & 13.4 & 23.9 & 28.9 \\
\flux + \ding       & 44.1 & 12.2 & 28.1 & 30.1 \\
\sd3 + ControlNet  & 45.8 & 17.0 & 27.0 & 24.4 \\
\sd3 + \ding        & 40.9 & 9.5 & 20.7 & 30.3 \\
\bottomrule
\end{tabular}
}
\label{tab:ding-vs-training-based}
\end{wraptable}
%
%
%
% --------humanedit--------
%  &  & FID & pFID & edFID & cPSNR \\
% denoiser & sampler &  &  &  &  \\
% \midrule
% flux & controlnet & 26.8 & 4.0 & 24.4 & 36.5 \\
% flux_fill & controlnet & 27.9 & 4.0 & 22.5 & 34.1 \\
% & ding & 31.1 & 4.9 & 28.1 & 35.9 \\
% sd3_medium & controlnet & 37.1 & 11.3 & 30.6 & 26.3 \\
%  & ding & 30.3 & 4.4 & 25.4 & 35.2 \\

% --------inpaintcoco--------
%  &  & FID & pFID & edFID & cPSNR \\
% denoiser & sampler &  &  &  &  \\
% \midrule
% flux & controlnet & 43.1 & 13.6 & 26.8 & 31.1 \\
% flux_fill & controlnet & 41.0 & 13.4 & 23.9 & 28.9 \\
% & ding & 44.1 & 12.2 & 28.1 & 30.1 \\
% sd3_medium & controlnet & 45.8 & 17.0 & 27.0 & 24.4 \\
%  & ding & 40.9 & 9.5 & 20.7 & 30.3 \\
%%
% \badr{the runtime of 10s corresponds to InpaintCOCO, but it is 30s for HumainEdit}
Modern large-scale diffusion models operate in a compressed latent space defined by an encoder--decoder pair $(\mathcal{E},\mathcal{D})$. While this design enables scalability to high-resolution data, it complicates training-free guidance, since inverse problems are defined in pixel space whereas guidance is applied in latent space, inducing a nonlinear likelihood through the decoder. For inpainting, however, \citet{avrahami2023blendedlatent} show that guidance can be performed entirely in latent space by downsampling the binary pixel mask and encoding the input as $x_*=\mathcal{E}(\mathrm{input})$. As illustrated in \Cref{fig:ding-inference-pipeline}, pixel- and latent-space masking are equivalent despite the encoder nonlinearity. This formulation enables inpainting directly in latent space;
% via linear problem,
% benefiting from reduced dimensionality and allowing the use of training-free methods restricted to linear inverse problems, 
particularly, the closed-form update in~\Cref{eq:ding-approx} is applicable.
We further discuss the limitations of this approach in \Cref{apdx:latent-inpainting}.

\paragraph{Experimental Setting.}
\begin{table}[t]
\captionsetup{font=small}
\vspace*{-8mm}
\caption{Quantitative comparison between training-free baselines on \humanedit\ 1024px and \inpaintcoco\ 512px on image editing tasks using \flux\ and \sd3 as priors. Runtime is limited to 50 NFEs.}
\vspace*{-2mm}
\resizebox{\textwidth}{!}{
% --------('humanedit', 'flux')--------
\begin{tabular}{l cccc | cccc | cccc | cccc}
% \begin{tabular}{l cccc c cccc}
\toprule
&  \multicolumn{4}{c}{\humanedit + \flux} &  \multicolumn{4}{c}{\humanedit + \sd3} &  \multicolumn{4}{c}{\inpaintcoco + \flux} &  \multicolumn{4}{c}{\inpaintcoco + \sd3} \\
\cmidrule(lr){2-5}  \cmidrule(lr){6-9} \cmidrule(lr){10-13} \cmidrule(lr){14-17}
& 
 FID & pFID & edFID & cPSNR &\, 
 FID & pFID & edFID & cPSNR &\, 
 FID & pFID & edFID & cPSNR &\, 
 FID & pFID & edFID & cPSNR \\
% sampler &  &  &  &  \\
\midrule
\blended & 30.8 & 4.1 & 25.1 & 34.6   &\, 33.1 & 8.9 & 30.6 & 31.5 &\, 41.9 & 10.4 & 23.3 & 28.0 &\, 46.0 & 14.1 & 27.6 & 27.2 \\ 
\ddnm & 31.2 & 4.8 & 31.4 & 36.4      &\, 30.5 & 4.6 & 27.7 & 34.6 &\, 45.6 & 14.2 & 32.9 & 30.4 &\, 41.6 & 11.6 & 26.2 & 29.8 \\
\diffpir & 31.1 & 4.3 & 26.0 & 35.4   &\, 31.1 & 4.6 & 26.4 & 33.8 &\, 43.6 & 11.9 & 26.9 & 28.9 &\, 42.2 & 10.0 & 22.3 & 28.5 \\
\ding & 31.2 & 5.0 & 27.5 & 35.7      &\, 30.8 & 4.7 & 25.5 & 34.4 &\, 43.2 & 11.9 & 26.9 & 29.9 &\, 41.9 & 10.3 & 21.9 & 29.2 \\
\flair & 30.9 & 5.2 & 27.1 & 35.0     &\, 30.7 & 4.9 & 31.4 & 34.2 &\, 41.5 & 15.7 & 33.1 & 29.2 &\, 42.7 & 11.8 & 29.9 & 29.3 \\
\flowchef & 32.3 & 5.7 & 29.4 & 33.5  &\, 34.6 & 9.0 & 27.9 & 31.3 &\, 52.2 & 17.0 & 40.9 & 29.5 &\, 46.6 & 16.2 & 28.2 & 28.0 \\
% 
%
% % --------('humanedit', 'flux')--------
% blended_diffusion & 30.8 & 4.1 & 25.1 & 34.6 \\
% ddnm & 31.2 & 4.8 & 31.4 & 36.4 \\
% diffpir & 31.1 & 4.3 & 26.0 & 35.4 \\
% ding & 31.2 & 5.0 & 27.5 & 35.7 \\
% flair & 30.9 & 5.2 & 27.1 & 35.0 \\
% flow_chef & 32.3 & 5.7 & 29.4 & 33.5 \\
%
% % --------('humanedit', 'sd3_medium')--------
% blended_diffusion & 33.1 & 8.9 & 30.6 & 31.5 \\
% ddnm & 30.5 & 4.6 & 27.7 & 34.6 \\
% diffpir & 31.1 & 4.6 & 26.4 & 33.8 \\
% ding & 30.8 & 4.7 & 25.5 & 34.4 \\
% flair & 30.7 & 4.9 & 31.4 & 34.2 \\
% flow_chef & 34.6 & 9.0 & 27.9 & 31.3 \\
%
% % --------('inpaintcoco', 'flux')--------
% blended_diffusion & 41.9 & 10.4 & 23.3 & 28.0 \\
% ddnm & 45.6 & 14.2 & 32.9 & 30.4 \\
% diffpir & 43.6 & 11.9 & 26.9 & 28.9 \\
% ding & 43.2 & 11.9 & 26.9 & 29.9 \\
% flair & 41.5 & 15.7 & 33.1 & 29.2 \\
% flow_chef & 52.2 & 17.0 & 40.9 & 29.5 \\
%
% % --------('inpaintcoco', 'sd3_medium')--------
% blended_diffusion & 46.0 & 14.1 & 27.6 & 27.2 \\
% ddnm & 41.6 & 11.6 & 26.2 & 29.8 \\
% diffpir & 42.2 & 10.0 & 22.3 & 28.5 \\
% ding & 41.9 & 10.3 & 21.9 & 29.2 \\
% flair & 42.7 & 11.8 & 29.9 & 29.3 \\
% flow_chef & 46.6 & 16.2 & 28.2 & 28.0 \\
%
\bottomrule
\end{tabular}
}
\vspace*{-4mm}
\label{tab:ding-vs-training-free}
\end{table}

Here, we describe the models, baselines, and datasets used in our experiments and defer the details to \Cref{apdx:models-baseslines}.
%For results, we highlight best \first{\transparent{0}{x}} and 2\textsuperscript{nd} best \second{\transparent{0}{x}}.

\emph{Large-scale text-conditioned priors.}\quad
For image editing, we consider, in addition to Stable Diffusion~3.5 (\sd3) which is exclusively used in \citet{moufad2025ding}, the large-scale text-to-image model \flux~\citep{batifol2025flux}.
We also extend the evaluation to video editing tasks: we adopt the text-to-video diffusion models \ltx~\citep{HaCohen2024LTX} and \wan~\citep{wan2025}.

\emph{Baselines.}\quad
We conduct benchmark using both training-free and training-based approaches. Among training-free baselines, we include
\blended~\citep{avrahami2023blendedlatent},
\diffpir~\citep{zhu2023denoising},
\ddnm~\citep{wang2023zeroshot},
\ding~\citep{moufad2025ding},
\flair~\citep{erbach2025solving},
and
\flowchef~\citep{patel2024flowchef}.
For training-based methods, we consider controlNet for inpainting methods~\citep{zhang2023controlnet} with \sd3\ and \flux\ as backbones.
We also include \flux\ Fill, a checkpoint specifically trained for image inpainting. For video inpainting, we compare with \wan-VACE, a checkpoint trained for video inpainting.

\emph{Datasets.}\quad%
\begin{wraptable}[17]{r}{0.28\textwidth}
\centering
\vspace*{-6mm}
\captionsetup{font=small}
\caption{Quantitative comparison on video editing tasks on \vpbench\ dataset
with resolution $(H, W, T)$=$(512,928,97)$
% For \ltx model, runtime is limited to XXX; for \wan\ to XXX.
All methods are training-free except last row.
Runtime is provided above the baselines.}
\vspace*{-2mm}
\resizebox{0.28\textwidth}{!}{
\begin{tabular}{l ccc}
\toprule
& CLIP & FVD & cPSNR \\
\cmidrule(lr){2-4}
\color{\runtimecolor}{2min 25s / video} &  \multicolumn{3}{c}{\ltx} \\
\midrule
\blended & 26.11 & 0.15 & 21.82 \\
\ddnm & 26.09 & 0.15 & 23.87 \\
\diffpir & 26.13 & 0.15 & 22.57 \\
\ding & 25.75 & 0.16 & 22.90 \\
\flair & 26.03 & 0.15 & 25.31 \\
\flowchef & 25.83 & 0.19 & 19.52 \\
\midrule
\\
\color{\runtimecolor}{5min 25s / video} &  \multicolumn{3}{c}{\wan} \\
% \cmidrule(lr){2-4}
\midrule
\blended & 26.33 & 0.14 & 26.95 \\
\ddnm & 26.31 & 0.13 & 30.00 \\
\diffpir & 26.36 & 0.13 & 28.52 \\
\ding & 26.24 & 0.13 & 29.21 \\
\flair & 26.30 & 0.13 & 29.94 \\
\flowchef & 26.17 & 0.16 & 25.22 \\
\cmidrule(lr){2-4}
\wan\-VACE & 26.29 & 0.11 & 32.69 \\
\bottomrule
\end{tabular}
}
\label{tab:video-results}
\end{wraptable}
% 
% 
% 
% 
% 
% --------('vpbench', 'ltx')--------
%  & CLIP & FVD & cPSNR \\
% sampler &  &  &  \\
% \midrule
% blended_diffusion & 26.11 & 0.15 & 21.82 \\
% ddnm & 26.09 & 0.15 & 23.87 \\
% diffpir & 26.13 & 0.15 & 22.57 \\
% ding & 25.75 & 0.16 & 22.90 \\
% flair & 26.03 & 0.15 & 25.31 \\
% flow_chef & 25.83 & 0.19 & 19.52 \\
% 
% 
% 
% --------('vpbench', 'wan')--------
%  & CLIP & FVD & cPSNR \\
% sampler &  &  &  \\
% \midrule
% blended_diffusion & 26.33 & 0.14 & 26.95 \\
% ddnm & 26.31 & 0.13 & 30.00 \\
% diffpir & 26.36 & 0.13 & 28.52 \\
% ding & 26.22 & 0.13 & 29.13 \\
% flair & 26.30 & 0.13 & 29.94 \\
% flow_chef & 26.17 & 0.16 & 25.22 \\
% vace & 26.29 & 0.11 & 32.69 \\%
We perform image editing experiments on \inpaintcoco~\citep{roesch2024inpaintcoco} and \humanedit~\citep{bai2024humanedit}.
The \inpaintcoco\ dataset contains 1{,}260 images, while we select a subset of 1{,}000 images from \humanedit. Both datasets provide editing masks paired with text prompts. For video editing, we use \vpbench~\citep{bai2024humanedit}, which consists of 133 videos, each annotated with editing masks and corresponding text prompts.

\paragraph{Evaluation.}
% \badr{le runtime pour ltx est de DING: 22it [02:25,  6.63s/it] et pour wan est DING: 22it [05:12, 14.22s/it]}
We consider a compute-constrained setting that mirrors deployment scenarios. When comparing training-free baselines among themselves, we fix the NFEs to $50$.
For comparisons between training-free and training-based methods, we instead match runtime to ensure a fair assessment; the runtimes are reported in the captions of each table.
Our evaluation primarily aims to further validate the approximation in \citet{moufad2025ding}, which we hence take as reference when benchmarking against training-based methods.
For image editing, we report FID and patch FID (pFID)~\citep{chai2022pfid}, the latter providing finer-grained evaluation for high-resolution images.
To measure preservation of the unedited content, we compute the context PSNR (cPSNR), defined as the PSNR over the unmasked regions only.
To assess the quality of the edited regions, we report the edited-region FID (edFID), obtained by extracting the edited regions and computing pFID over them. For video editing, we report CLIP-Score~\citep{radford2021learning}, FVD~\citep{unterthiner2018fvd}, and cPSNR.
For FID related metrics, lower is better, for CLIP and cPSNR higher is better.

\emph{Comments.}\quad
Quantitative results are summarized in \Cref{tab:ding-vs-training-free,tab:ding-vs-training-based,tab:video-results}, with qualitative comparisons shown in \Cref{fig:ding_vs_trained_sd3,fig:ding_vs_trained_flux_1,fig:ding_vs_trained_flux_2}. Training-free baselines achieve competitive performance across metrics despite not being trained for such tasks. In particular, \Cref{tab:ding-vs-training-based} shows that, under the same compute budgets, the training-free method \ding\ performs on par with, and in even surpasse the training-based method \sd3\ + ControlNet. Qualitative examples in \Cref{fig:ding_vs_trained_sd3,fig:ding_vs_trained_flux_1,fig:ding_vs_trained_flux_2} further corroborate these findings. It shows high-quality edits and strong controllability for training-free methods, with visual performance comparable to training-based counterparts.
Qualitative results on video editing tasks can be found on the project webpage\footnote{\url{\webpageURL}}.

\paragraph{Python package for editing.}
We release \packageName, a Python package for training-free editing via inpainting that implements the pipeline in~\Cref{fig:ding-inference-pipeline}. It currently supports image and video editing with three image priors and two video priors; audio support is under active development. \packageName\ is model- and sampler-agnostic: integration of new diffusion priors and training-free samplers is straightforward. The package also provides evaluation scripts, pre-implemented image and video metrics, and comprehensive installation and usage documentation to facilitate reproducibility.

% We release \packageName, a Python package for editing via inpainting with training-free methods. The package implements the editing pipeline in~\cref{fig:ding-inference-pipeline} and currently supports three modalities: images, videos, and audio. \packageName\ includes three image priors and two video priors. Support for audio is under active development and will be released shortly.
% %
% The design of \packageName\ is model and sampler-agnostic.
% allowing it to interface seamlessly with virtually any training-free samplers and diffusion priors.
% This architecture facilitates the straightforward integration of new samplers and models as they become available. In addition, the package includes evaluation scripts for testing and benchmarking, along with pre-implemented metrics for both image and video modalities. Step-by-step installation and usage guides are provided to support reproducibility and ease of adoption.

% \begin{wrapfigure}{r}{0.3\textwidth}
%     % \vspace*{-3mm}
%     \centering
%     \includegraphics[width=0.3\textwidth]{figures/ding_package_light.png}
%     \captionsetup{font=small}
%     % \vspace*{-3mm}
%     \caption{XXX}
%     \label{fig:ding-package}
% \end{wrapfigure}

% \section{Conclusion}

\section*{Acknowledgements}
The work of Ahmed Ghorbel and Badr Moufad has been supported by Technology Innovation Institute (TII), project Fed2Learn. The work of Eric Moulines has been partly funded by the European Union (ERC-2022-SYG-OCEAN-101071601). Views and opinions expressed are however those
of the author(s) only and do not necessarily reflect those of the European Union or the European Research Council Executive Agency. Neither the European Union nor the granting authority can be held responsible for them. This work was granted access to the HPC resources of IDRIS under the allocations 2025-AD011016497 by GENCI.

\clearpage
\newpage
% --- biblio
\bibliographystyle{iclr2026_conference}
\bibliography{bibliography}

@article{avrahami2023blendedlatent,
        author = {Avrahami, Omri and Fried, Ohad and Lischinski, Dani},
        title = {Blended Latent Diffusion},
        year = {2023},
        issue_date = {August 2023},
        publisher = {Association for Computing Machinery},
        address = {New York, NY, USA},
        volume = {42},
        number = {4},
        issn = {0730-0301},
        url = {https://doi.org/10.1145/3592450},
        doi = {10.1145/3592450},
        journal = {ACM Trans. Graph.},
        month = {jul},
        articleno = {149},
        numpages = {11}
}

@article{albergo2023stochastic,
  title = {Stochastic interpolants: A unifying framework for flows and diffusions},
  author = {Albergo, Michael S and Boffi, Nicholas M and Vanden-Eijnden, Eric},
  journal = {arXiv preprint arXiv:2303.08797},
  year = {2023}
}

@article{boys2023tweedie,
  title = {Tweedie moment projected diffusions for inverse problems},
  author = {Boys, Benjamin and Girolami, Mark and Pidstrigach, Jakiw and Reich, Sebastian and Mosca, Alan and Akyildiz, O Deniz},
  journal = {arXiv preprint arXiv:2310.06721},
  year = {2023}
}

@inproceedings{lipman2023flow,
  title = {Flow Matching for Generative Modeling},
  author = {Yaron Lipman and Ricky T. Q. Chen and Heli Ben-Hamu and Maximilian Nickel and Matthew Le},
  booktitle = {The Eleventh International Conference on Learning Representations },
  year = {2023},
  url = {https://openreview.net/forum?id=PqvMRDCJT9t}
}

@article{batifol2025flux,
  title = {FLUX. 1 Kontext: Flow Matching for In-Context Image Generation and Editing in Latent Space},
  author = {Batifol, Stephen and Blattmann, Andreas and Boesel, Frederic and Consul, Saksham and Diagne, Cyril and Dockhorn, Tim and English, Jack and English, Zion and Esser, Patrick and Kulal, Sumith and others},
  journal = {arXiv e-prints},
  pages = {arXiv--2506},
  year = {2025}
}

@article{wu2025qwen,
  title = {Qwen-image technical report},
  author = {Wu, Chenfei and Li, Jiahao and Zhou, Jingren and Lin, Junyang and Gao, Kaiyuan and Yan, Kun and Yin, Sheng-ming and Bai, Shuai and Xu, Xiao and Chen, Yilei and others},
  journal = {arXiv preprint arXiv:2508.02324},
  year = {2025}
}

@article{erbach2025solving,
  title = {Solving Inverse Problems with FLAIR},
  author = {Erbach, Julius and Narnhofer, Dominik and Dombos, Andreas and Schiele, Bernt and Lenssen, Jan Eric and Schindler, Konrad},
  journal = {arXiv preprint arXiv:2506.02680},
  year = {2025}
}

@article{daras2024survey,
  title = {A survey on diffusion models for inverse problems},
  author = {Daras, Giannis and Chung, Hyungjin and Lai, Chieh-Hsin and Mitsufuji, Yuki and Ye, Jong Chul and Milanfar, Peyman and Dimakis, Alexandros G and Delbracio, Mauricio},
  journal = {arXiv preprint arXiv:2410.00083},
  year = {2024}
}

@inproceedings{esser2024scaling,
  title = {Scaling rectified flow transformers for high-resolution image synthesis},
  author = {Esser, Patrick and Kulal, Sumith and Blattmann, Andreas and Entezari, Rahim and M{\"u}ller, Jonas and Saini, Harry and Levi, Yam and Lorenz, Dominik and Sauer, Axel and Boesel, Frederic and others},
  booktitle = {Forty-first international conference on machine learning},
  year = {2024}
}

@article{karras2022elucidating,
  title = {Elucidating the design space of diffusion-based generative models},
  author = {Karras, Tero and Aittala, Miika and Aila, Timo and Laine, Samuli},
  journal = {Advances in Neural Information Processing Systems},
  volume = {35},
  pages = {26565--26577},
  year = {2022}
}

@inproceedings{rezende2015variational,
  title = {Variational inference with normalizing flows},
  author = {Rezende, Danilo and Mohamed, Shakir},
  booktitle = {International conference on machine learning},
  pages = {1530--1538},
  year = {2015},
  organization = {PMLR}
}

@inproceedings{zhu2023denoising,
  title = {Denoising diffusion models for plug-and-play image restoration},
  author = {Zhu, Yuanzhi and Zhang, Kai and Liang, Jingyun and Cao, Jiezhang and Wen, Bihan and Timofte, Radu and Van Gool, Luc},
  booktitle = {Proceedings of the IEEE/CVF Conference on Computer Vision and Pattern Recognition},
  pages = {1219--1229},
  year = {2023}
}

@inproceedings{wang2023zeroshot,
  title = {Zero-Shot Image Restoration Using Denoising Diffusion Null-Space Model},
  author = {Yinhuai Wang and Jiwen Yu and Jian Zhang},
  booktitle = {The Eleventh International Conference on Learning Representations },
  year = {2023},
  url = {https://openreview.net/forum?id=mRieQgMtNTQ}
}

@article{song2019generative,
  title = {Generative modeling by estimating gradients of the data distribution},
  author = {Song, Yang and Ermon, Stefano},
  journal = {Advances in neural information processing systems},
  volume = {32},
  year = {2019}
}

@article{patel2024flowchef,
  title = {Steering Rectified Flow Models in the Vector Field for Controlled Image Generation},
  author = {Patel, Maitreya and Wen, Song and Metaxas, Dimitris N. and Yang, Yezhou},
  journal = {arXiv preprint arXiv:2412.00100},
  year = {2024}
}

@inproceedings{chung2023diffusion,
  title = {Diffusion Posterior Sampling for General Noisy Inverse Problems},
  author = {Hyungjin Chung and Jeongsol Kim and Michael Thompson Mccann and Marc Louis Klasky and Jong Chul Ye},
  booktitle = {The Eleventh International Conference on Learning Representations },
  year = {2023},
  url = {https://openreview.net/forum?id=OnD9zGAGT0k}
}

@article{ho2020denoising,
  title = {Denoising diffusion probabilistic models},
  author = {Ho, Jonathan and Jain, Ajay and Abbeel, Pieter},
  journal = {Advances in Neural Information Processing Systems},
  volume = {33},
  pages = {6840--6851},
  year = {2020}
}

@article{kawar2021snips,
  title = {SNIPS: Solving noisy inverse problems stochastically},
  author = {Kawar, Bahjat and Vaksman, Gregory and Elad, Michael},
  journal = {Advances in Neural Information Processing Systems},
  volume = {34},
  pages = {21757--21769},
  year = {2021}
}

@article{kadkhodaie2020solving,
  title = {Solving linear inverse problems using the prior implicit in a denoiser},
  author = {Kadkhodaie, Zahra and Simoncelli, Eero P},
  journal = {arXiv preprint arXiv:2007.13640},
  year = {2020}
}

@inproceedings{song2021ddim,
  title = {Denoising Diffusion Implicit Models},
  author = {Jiaming Song and Chenlin Meng and Stefano Ermon},
  booktitle = {International Conference on Learning Representations},
  year = {2021},
  url = {https://openreview.net/forum?id=St1giarCHLP}
}

@article{roberts:tweedie:1996,
  author = {Roberts, G. O. and Tweedie, R. L.},
  date = {1996-03},
  journaltitle = {Biometrika},
  title = {{Geometric convergence and central limit theorems for multidimensional {H}astings and {M}etropolis algorithms}},
  doi = {10.1093/biomet/83.1.95},
  eprint = {https://academic.oup.com/biomet/article-pdf/83/1/95/709644/83-1-95.pdf},
  issn = {0006-3444},
  number = {1},
  pages = {95-110},
  url = {https://doi.org/10.1093/biomet/83.1.95},
  volume = {83}
}

@book{bishop2006pattern,
  author = {Bishop, Christopher M.},
  title = {Pattern Recognition and Machine Learning (Information Science and Statistics)},
  year = {2006},
  isbn = {0387310738},
  publisher = {Springer-Verlag},
  address = {Berlin, Heidelberg}
}

@inproceedings{mardani2024a,
  title = {A Variational Perspective on Solving Inverse Problems with Diffusion Models},
  author = {Morteza Mardani and Jiaming Song and Jan Kautz and Arash Vahdat},
  booktitle = {The Twelfth International Conference on Learning Representations},
  year = {2024},
  url = {https://openreview.net/forum?id=1YO4EE3SPB}
}

@inproceedings{song2021score,
  author = {Song, Yang and Sohl-Dickstein, Jascha and Kingma, Diederik P and Kumar, Abhishek and Ermon, Stefano and Poole, Ben},
  booktitle = {International Conference on Learning Representations},
  title = {Score-Based Generative Modeling through Stochastic Differential Equations},
  year = {2021}
}

@article{janati2025mgdm,
  title = {A Mixture-Based Framework for Guiding Diffusion Models},
  author = {Janati, Yazid and Moufad, Badr and Abou El Qassim, Mehdi and
            Durmus, Alain and Moulines, Eric and Olsson, Jimmy},
  journal = {preprint},
  year = {2025}
}

@inproceedings{chai2022pfid,
  title = {Any-resolution training for high-resolution image synthesis},
  author = {Chai, Lucy and Gharbi, Michael and Shechtman, Eli and Isola, Phillip and Zhang, Richard},
  booktitle = {European conference on computer vision},
  pages = {170--188},
  year = {2022},
  organization = {Springer}
}

@inproceedings{radford2021learning,
  title = {Learning transferable visual models from natural language supervision},
  author = {Radford, Alec and Kim, Jong Wook and Hallacy, Chris and Ramesh, Aditya and Goh, Gabriel and Agarwal, Sandhini and Sastry, Girish and Askell, Amanda and Mishkin, Pamela and Clark, Jack and others},
  booktitle = {International conference on machine learning},
  pages = {8748--8763},
  year = {2021},
  organization = {PmLR}
}

@article{moufad2025ding,
  title={Efficient Zero-Shot Inpainting with Decoupled Diffusion Guidance},
  author={Moufad, Badr and Shouraki, Navid Bagheri and Durmus, Alain Oliviero and Hirtz, Thomas and Moulines, Eric and Olsson, Jimmy and Janati, Yazid},
  journal={arXiv preprint arXiv:2512.18365},
  year={2025}
}

@article{chen2018neural-ode,
  title={Neural ordinary differential equations},
  author={Chen, Ricky TQ and Rubanova, Yulia and Bettencourt, Jesse and Duvenaud, David K},
  journal={Advances in neural information processing systems},
  volume={31},
  year={2018}
}

@article{wan2025,
      title={Wan: Open and Advanced Large-Scale Video Generative Models}, 
      author={Team Wan and Ang Wang and Baole Ai and Bin Wen and Chaojie Mao and Chen-Wei Xie and Di Chen and Feiwu Yu and Haiming Zhao and Jianxiao Yang and Jianyuan Zeng and Jiayu Wang and Jingfeng Zhang and Jingren Zhou and Jinkai Wang and Jixuan Chen and Kai Zhu and Kang Zhao and Keyu Yan and Lianghua Huang and Mengyang Feng and Ningyi Zhang and Pandeng Li and Pingyu Wu and Ruihang Chu and Ruili Feng and Shiwei Zhang and Siyang Sun and Tao Fang and Tianxing Wang and Tianyi Gui and Tingyu Weng and Tong Shen and Wei Lin and Wei Wang and Wei Wang and Wenmeng Zhou and Wente Wang and Wenting Shen and Wenyuan Yu and Xianzhong Shi and Xiaoming Huang and Xin Xu and Yan Kou and Yangyu Lv and Yifei Li and Yijing Liu and Yiming Wang and Yingya Zhang and Yitong Huang and Yong Li and You Wu and Yu Liu and Yulin Pan and Yun Zheng and Yuntao Hong and Yupeng Shi and Yutong Feng and Zeyinzi Jiang and Zhen Han and Zhi-Fan Wu and Ziyu Liu},
      journal = {arXiv preprint arXiv:2503.20314},
      year={2025}
}

@inproceedings{zhang2023controlnet,
  title={Adding conditional control to text-to-image diffusion models},
  author={Zhang, Lvmin and Rao, Anyi and Agrawala, Maneesh},
  booktitle={Proceedings of the IEEE/CVF international conference on computer vision},
  pages={3836--3847},
  year={2023}
}

@article{HaCohen2024LTX,
  title={LTX-Video: Realtime Video Latent Diffusion},
  author={HaCohen, Yoav and Chiprut, Nisan and Brazowski, Benny and Shalem, Daniel and Moshe, Dudu and Richardson, Eitan and Levin, Eran and Shiran, Guy and Zabari, Nir and Gordon, Ori and Panet, Poriya and Weissbuch, Sapir and Kulikov, Victor and Bitterman, Yaki and Melumian, Zeev and Bibi, Ofir},
  journal={arXiv preprint arXiv:2501.00103},
  year={2024}
}

@misc{roesch2024inpaintcoco,
      title={Enhancing Conceptual Understanding in Multimodal Contrastive Learning through Hard Negative Samples}, 
      author={Philipp J. Rösch and Norbert Oswald and Michaela Geierhos and Jindřich Libovický},
      year={2024},
      eprint={2403.02875},
      archivePrefix={arXiv},
      primaryClass={cs.CV}
}

@article{bai2024humanedit,
  title={HumanEdit: A High-Quality Human-Rewarded Dataset for Instruction-based Image Editing},
  author={Bai, Jinbin and Chow, Wei and Yang, Ling and Li, Xiangtai and Li, Juncheng and Zhang, Hanwang and Yan, Shuicheng},
  journal={arXiv preprint arXiv:2412.04280},
  year={2024}
}

@misc{flux-2-2025,
    author={Black Forest Labs},
    title={{FLUX.2: Frontier Visual Intelligence}},
    year={2025},
    howpublished={\url{https://bfl.ai/blog/flux-2}},
}

@article{unterthiner2018fvd,
  title={Towards accurate generative models of video: A new metric \& challenges},
  author={Unterthiner, Thomas and Van Steenkiste, Sjoerd and Kurach, Karol and Marinier, Raphael and Michalski, Marcin and Gelly, Sylvain},
  journal={arXiv preprint arXiv:1812.01717},
  year={2018}
}

% --- appendix
\clearpage
\newpage

\paragraph{Reproducibility statement.}
All experiments are reproducible using the released Python package \packageName\footnote{\url{\packageURL}}, which includes implementations of all training-free and training-based methods considered in this work, along with the corresponding evaluation scripts.

\paragraph{Ethics statement.}
Test-time guidance lowers computational barriers for creative media applications, broadening accessibility. However, diffusion-based inpainting methods are inherently dual-use and may be misused to generate deceptive and harmful content, such as manipulated images or synthetic media that obscure authenticity. This risk underscores the importance of responsible deployment, including clear usage guidelines and traceable outputs.

% \paragraph{Reproducibility statement.} All experiments can be reproduced using the released Python package \packageName\footnote{\url{\packageURL}}. All considered training-based and training-free methods are provided and evaluation scripts also are provided.

% \paragraph{Ethics statement.} We demonstrates that test-time guidance provide a clear benefits for applications in creative media.
% This mainly lowers the barriers for adoption namely in terms of computation.
% However rises ethical considerations. Diffusion-based inpainting methods can be misappropriated for producing deceptive and harmful content, such as manipulated images or synthetic media that obscure authenticity. This dual-use nature highlights the need for proactive safeguards, including transparent usage guidelines, traceable model outputs, and continued development of forensic detection tools to ensure responsible integration of such technologies.

\appendix
% \section{Models and baselines details}
% \label{apdx:models-baseslines}

\section{Details on models and baselines}
\label{apdx:models-baseslines}

\paragraph{Large-scale models.}
For image editing, we use \sd3~\citep{esser2024scaling} and \flux~\citep{batifol2025flux}, both equipped with a VAE featuring a spatial downsampling factor of $\times 8$. For video editing, we consider \ltx~\citep{HaCohen2024LTX}, which employs a video VAE with spatial and temporal downsampling factors of $\times 32$ and $\times 8$, respectively, and \wan, which uses $\times 8$ spatial and $\times 3$ temporal downsampling. For all models, we use the default guidance scale provided by the authors.

\paragraph{Training-based methods.}
We include baselines that are explicitly trained for inpainting. For images, we evaluate ControlNet-based approaches, including \sd3\ ControlNet\footnote{\url{huggingface.co/alimama-creative/SD3-Controlnet-Inpainting}} and \flux\ ControlNet\footnote{\url{huggingface.co/alimama-creative/FLUX.1-dev-Controlnet-Inpainting-Beta}}. The \sd3\ ControlNet model is fine-tuned on a large-scale dataset of 12 million image--mask pairs at $1024 \times 1024$ resolution. Similarly, \flux\ ControlNet is trained on 15 million images.

We also include \flux\ Fill, a \flux-based checkpoint trained specifically for inpainting\footnote{\url{huggingface.co/black-forest-labs/FLUX.1-Fill-dev}}. For video inpainting, we evaluate \wan-VACE~1.3B, a dedicated video inpainting checkpoint.

\paragraph{Training-free baselines.}
We adopt the hyperparameters reported in \citet[Appendix Table~8]{moufad2025ding}. As \flair~\citep{erbach2025solving} was not considered in \citet{moufad2025ding}, we detail below its implementation and hyperparameters choices.
We follow \citet[Algorithm~1]{erbach2025solving} and build upon their official codebase\footnote{\url{https://github.com/prs-eth/FLAIR/}}. 
We set the number of inner likelihood optimization steps to \texttt{n\_likelihood\_steps}$=15$ with early stopping threshold \texttt{early\_stopping}$=10^{-4}$. Since we consider different prior models from those in the original work, namely \sd3, we use a fixed regularization weight of $1$, which we found to perform robustly in practice without requiring calibration from the pre-computed flow-matching loss.

\section{Inpainting in latent space}
\label{apdx:latent-inpainting}

% Most modern large-scale diffusion models for images and videos operate in a compressed latent space defined by an encoder--decoder pair $(\mathcal{E}, \mathcal{D})$
A key observation of \citet{avrahami2023blendedlatent} is that, for the case of inpainting, guidance can be carried out entirely in latent space by deriving an appropriate latent mask: the binary pixel space mask is downsampled according; while The input is directly encoded as $x_* = \mathcal{E}(\mathrm{input})$.
% As illustrated in \Cref{fig:ding-inference-pipeline}, masking in pixel space is equivalent to masking in latent space despite the nonlinearity of the encoder.
This equivalence allows inpainting problems to be formulated directly in latent space and benefit from the its compressed representations.
Moreover, it enables the application of training-free guidance methods that are valid only for linear inverse problems; in particular, the closed-form update in \Cref{eq:ding-approx} applies in this setting.

% Most modern large-scale diffusion models for images and videos operate in a compressed latent space defined by an encoder--decoder pair $(\mathcal{E}, \mathcal{D})$.
% The compression involves a spatial/temporal downsampling factors. While this enables diffusion models to scale to high-resolution inputs, its complicates the use of training-free guidance methods. In particular, as inverse problems are defined in pixel space whereas guidance is performed in latent space, the resulting likelihood involves a nonlinear composition with the decoder $\mathcal{D}$, which make inference both conceptually and computationally challenging.
%
% A key observation of \citet{avrahami2023blendedlatent} is that, for the case of inpainting, guidance can be carried out entirely in latent space by deriving an appropriate latent mask: the binary pixel space mask is downsampled according; while The input is directly encoded as $x_* = \mathcal{E}(\mathrm{input})$.
% As illustrated in \Cref{fig:ding-inference-pipeline}, masking in pixel space is equivalent to masking in latent space despite the nonlinearity of the encoder.
% This equivalence allows inpainting problems to be formulated directly in latent space and benefit from the compressed latent space. Moreover, it enables the application of training-free guidance methods that are valid only for linear inverse problems; in particular, the closed-form update \eqref{eq:ding-approx} applies in this setting.
% A complete description of the editing-via-inpainting pipeline is provided in \Cref{fig:ding-inference-pipeline}.

\begin{figure}
    \centering
    \includegraphics[width=\linewidth]{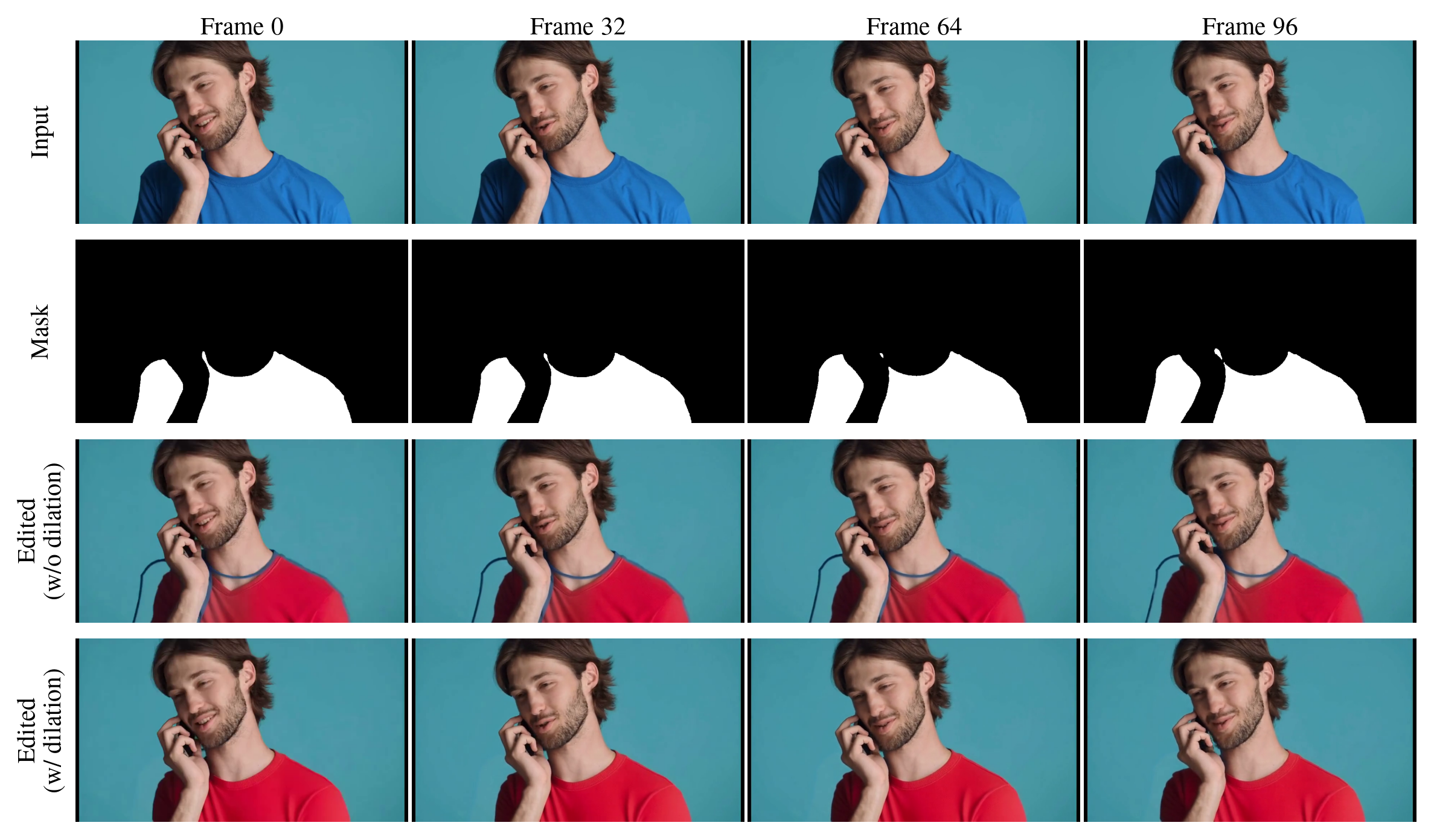}
    \caption{Visualization of \emph{context leakage} in latent-space video inpainting. When lifting a pixel-space inpainting task to the latent space, downsampling the mask without adjustment can lead to boundary artifacts. From top  to bottom row: input video, binary edit masks, reconstruction using naive downsampled masks, and reconstruction using the dilated masks. Note that naive downsampling (third row) causes the t-shirt's original boundary (blue outline) to leak into the latent reconstruction, whereas dilation (fourth row) successfully avoid this issue.}
    \label{fig:limitation-mask-latent}
\end{figure}

\emph{Limitations.}\quad
Defining inpainting directly in latent space entails several limitations, as noted by \citet{avrahami2023blendedlatent}. First, the quality of the reconstruction is inherently constrained by the encoder--decoder pair of the diffusion model, although this limitation becomes less pronounced as recent models continue to improve \citep{flux-2-2025}.
Second, the encoder's downsampling factor limits the spatial granularity of masks that can be represented: masked regions smaller than the downsampling scale may be ignored, leading to ineffective or inaccurate edits. Additionally, small or thin masked regions can introduce artifacts due to leakage from unmasked regions, since the downsampled mask may not fully cover the intended area.
\Cref{fig:limitation-mask-latent} illustrates this limitation in practice: the outline of the blue shirt in the input video leaks into the edited result. The effect becomes more pronounced as the encoder's downsampling factor increases. In particular, we noticed in practice that this issue is more evident for the \ltx\ video model, which employs a spatial downsampling factor of $\times 32$, compared to \wan, which uses a factor of $\times 8$.
We mitigated this issue by slightly overestimating the support of the mask near the boundaries as illustrated in \Cref{fig:ding-inference-pipeline},
which alleviates boundary artifacts at the cost of slightly altering the observed regions around the mask edges.% as we further illustrate in Fig.~XXX.
% which alleviates boundary artifacts as we further illustrate in Fig.~XXX.

\section{Examples of editing via inpainting}
\label{apdx:reconstructions}
Here, we provide a side-by-side comparison of the \ding\ and the considered baselines on image editing tasks via inpainting on \humanedit.
The red overlay in the first column shows the masked region to be edited and the text in the left-hand side of each row represents the editing prompt.

Qualitative results on video editing tasks can be found on the project webpage\footnote{\url{\webpageURL}}.

\vspace*{5mm}
\emph{(See the next pages for the gallery of examples)}
\vspace*{5mm}

\begin{figure}[t]
    \centering
    \includegraphics[width=0.48\textwidth,keepaspectratio]{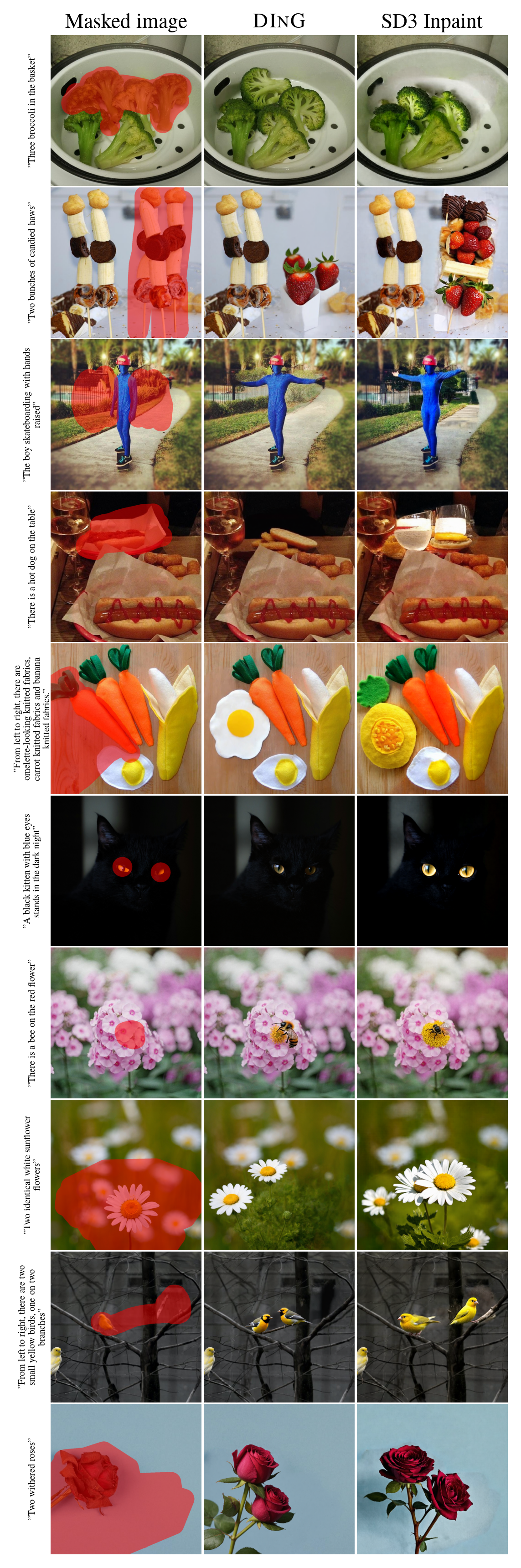}%
    \hfill%
    \includegraphics[width=0.48\textwidth,keepaspectratio]{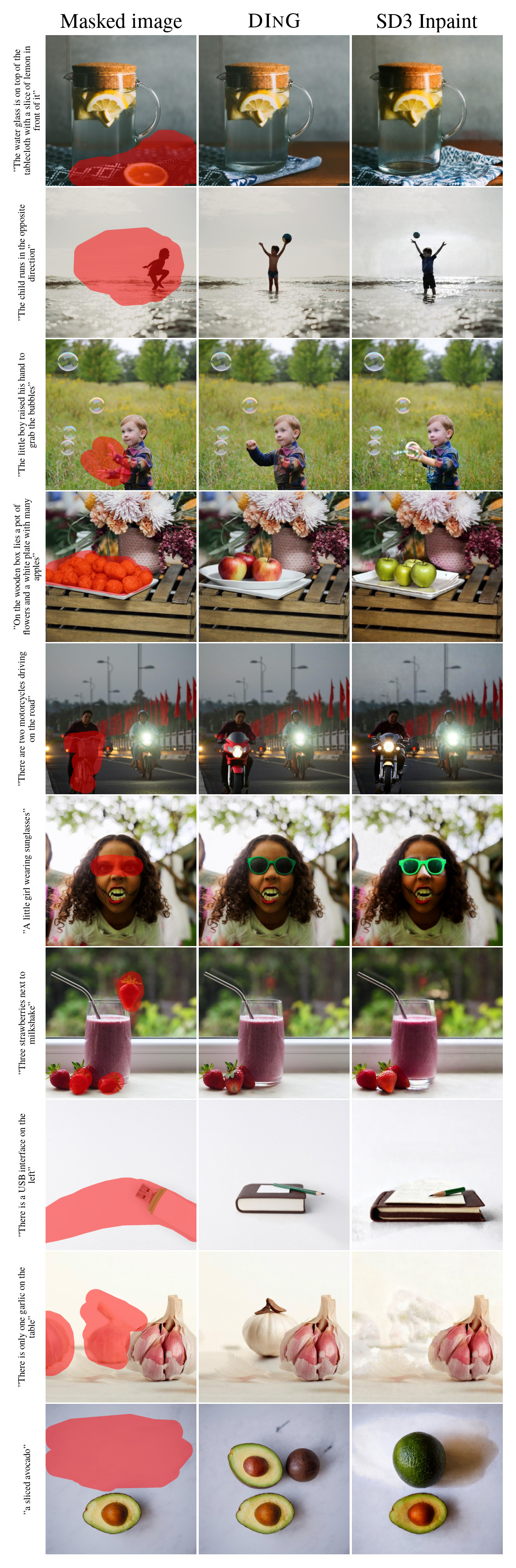}%
    \caption{Qualitative comparison of \ding\ and SD3 with ControlNet (SD3 Inpaint) on \humanedit. The methods are limited to a runtime of 30 seconds per image}
    \label{fig:ding_vs_trained_sd3}
\end{figure}

\clearpage
\begin{figure}
    \centering
    \includegraphics[width=0.75\linewidth]{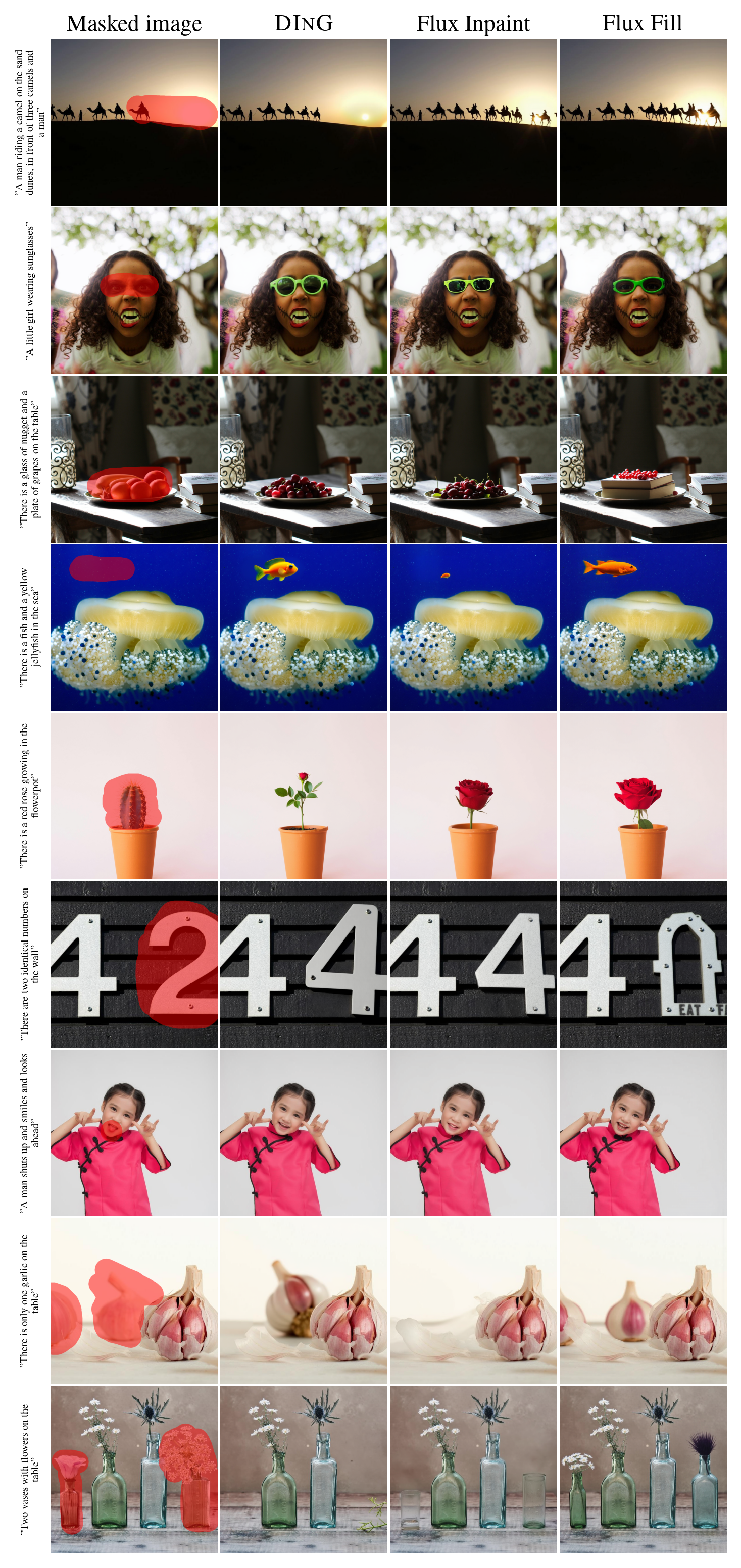}
    \caption{Qualitative comparison 1 of \ding\, Flux with ControlNet (Flux Inpaint), and Flux Fill on \humanedit. The methods are limited to a runtime of 30 seconds per image.}
    \label{fig:ding_vs_trained_flux_1}
\end{figure}

\clearpage
\begin{figure}
    \centering
    \includegraphics[width=\linewidth]{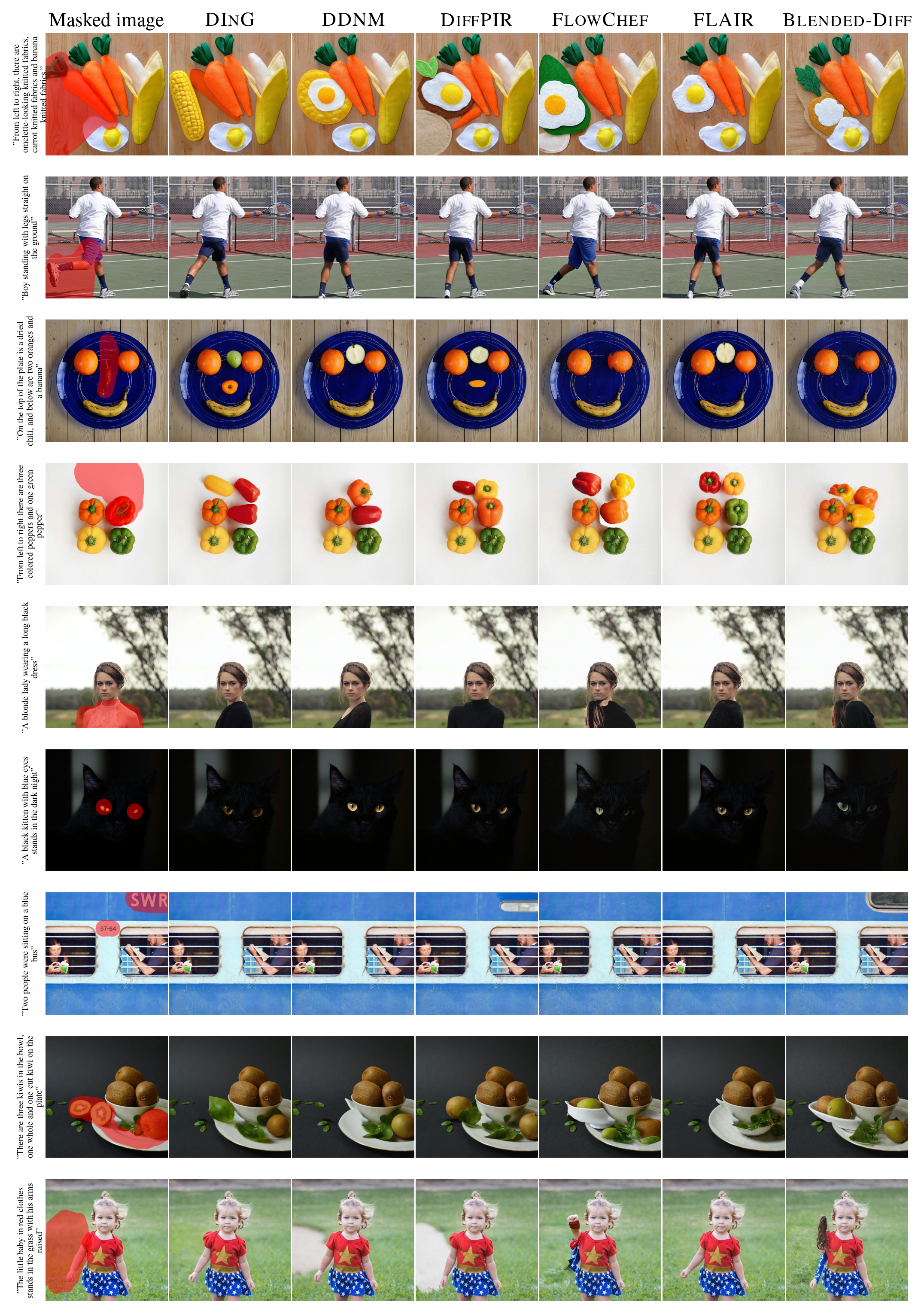}
    \caption{Qualitative comparison between and training-free baselines on \humanedit\ with \sd3 model as prior. The methods are limited to NFE 50.}
    \label{fig:ding_vs_trained_flux_2}
\end{figure}

\end{document}